\documentclass[fleqn,10pt]{JLA_article} 

\usepackage[english]{babel} 

\usepackage{hyperref} 
\hypersetup{hidelinks,colorlinks,breaklinks=true,urlcolor=color2,citecolor=color1,linkcolor=color1,bookmarksopen=false,pdftitle={Agreement Is Not Validity: Cross-Model LLM Consensus in Diagnosing Student Failure Modes in K–12 Math Tutoring Dialogue},pdfauthor={Clayton Cohn, Joyce Fonteles, Kirk Vanacore, Gianni Mazza, Candida Crawford, Tom Hooper, Gautam Biswas, and Rene Kizilcec}}
\usepackage{subcaption} 
\usepackage{wrapfig}
\usepackage{placeins}

\PaperTitle{Agreement Is Not Validity: Cross-Model LLM Consensus in Diagnosing Student Failure Modes in K–12 Math Tutoring Dialogue} 

\Authors{Clayton Cohn\textsuperscript{1}*, Joyce Fonteles\textsuperscript{2}, Kirk Vanacore\textsuperscript{1}, Gianni Mazza\textsuperscript{3}, Candida Crawford\textsuperscript{3}, Tom Hooper\textsuperscript{3}, Gautam Biswas\textsuperscript{2}, and Rene Kizilcec\textsuperscript{1}} 
\affiliation{\textsuperscript{1}\textit{College of Computing and Information Science, Cornell University, Ithaca, USA.}} 
\affiliation{\textsuperscript{2}\textit{College of Connected Computing, Vanderbilt University, Nashville, USA.}} 
\affiliation{\textsuperscript{3}\textit{Third Space Learning, Swindon, UK.}} 
\affiliation{\textit{*Corresponding author: clayton.cohn@cornell.edu}} 


\Submitted{9/28/26}
\Accepted{}
\Published{}

\Volume{1}
\Number{1}
\Pages{1---10}
\Doi{xxx-xxx-xxx}
\Notesname{Notes for Practice} 

\note{Agreement among multiple LLMs should not be treated as evidence that a learner diagnosis is correct: models can reach strong consensus while sharing systematic disagreements with the human reference.}

\note{LLM-generated diagnoses of student difficulty should be independently validated against human judgment before being used to inform tutoring decisions, learner models, or downstream learning analytics.}

\note{LLMs may over-diagnose failure modes when there is no available evidence to support diagnosis; automated tutoring systems should preserve the option to remain uncertain rather than force a classification.}

\note{When trained experts show only modest agreement, neither human nor LLM judgments should be treated as ground truth; consequential uses should include adjudication rather than unadjudicated classification.}

\Abstract{In K–12 mathematics tutoring, student–tutor dialogue provides rich evidence of learners’ problem-solving processes and sources of difficulty. Learning analytics research increasingly relies on large language models (LLMs) to extract such information from dialogue for a variety of downstream tasks, including knowledge tracing, behavioral modeling, and diagnosis of student reasoning errors. However, the validity of these model-generated interpretations remains insufficiently understood. In this exploratory study, we examine the validity of LLM classifications of five student failure modes in mathematics tutoring dialogue using an operational diagnostic codebook: uncertainty, misattribution, operator selection, conceptual gap, and procedural slip. Across models, human--LLM agreement was moderate ($\kappa=.524$--$.597$), while cross-model agreement was substantially higher ($\kappa=.755$--$.781$; $\alpha=.769$). These findings show that cross-model agreement can create a misleading appearance of correctness, challenging the assumption that consensus among LLMs constitutes evidence of valid learner interpretation. For learning analytics, the implication is clear: scalable labeling is useful only if the inferred constructs are valid, and model consensus cannot substitute for independent evidence of that validity.\\}

\begin{document}

\flushbottom 

\maketitle 


\thispagestyle{fancy} 




\section{Introduction and Related Work}

\subsection{Theoretical Grounding and Student Discourse}

Human tutoring has long been shown to support mathematics learning among K--12 students \cite{borchers2026brief}, creating opportunities for sustained, responsive interaction around students' emerging understanding. The ICAP (Interactive--Constructive--Active--Passive) framework explains why such interaction matters, proposing that learning deepens as students move toward constructive and interactive engagement \cite{chi2014icap}. This perspective positions student discourse as rich evidence from which learning analytics systems can infer properties of learners' understanding and difficulty.

Characterizing student struggle is paramount. In the United States, for example, the Education Scorecard describes mathematics achievement below levels observed a decade earlier in 70\% of school districts \cite{dewey2026learning}. Addressing such declines requires understanding not only whether students are struggling, but the nature of that struggle as it unfolds. Student--tutor discourse can reveal distinct \emph{failure modes}: the underlying reasons a student produces an incorrect response or cannot complete a learning task. Distinguishing these modes is critical, as different sources of difficulty often call for different instructional responses. Traditionally, doing so required trained human annotators to interpret discourse manually and label instances of struggle---a time- and resource-intensive process that limits analysis at scale \cite{garg2024automated}.

\subsection{LLM Labeling in Learning Analytics}

Large language models (LLMs) have consequently become attractive for transforming conversational learning traces into scalable representations of learners' understanding and difficulty. Recent work has used LLMs to code tutoring and classroom discourse \cite{garg2024automated}, generate labels supporting knowledge tracing \cite{scarlatos2025exploring}, and conduct qualitative coding across educational contexts \cite{liu2025qualitative}; while other approaches retain humans in the loop \cite{xavier2026solo}. However, evidence that LLMs can replicate human annotation remains mixed; \citeA{vanacore2026well} conclude that ``unscaffolded LLMs are not yet sufficiently reliable to replace human coders for complex instructional constructs.'' 

Learning analytics researchers have therefore begun to scrutinize how such annotation should be evaluated and improved. \citeA{xu_enhancing_2026} distinguish task ambiguity from model-specific annotation error; \citeA{borchers2026learning} show that conventional reliability estimates may inadequately characterize LLM scoring in longitudinal learning settings; and \citeA{ahtisham2026ai} use self- and cross-model verification to improve LLM coding of tutoring discourse. Together, these studies demonstrate the promise of LLMs for scalable annotation while underscoring the importance of validating whether their labels faithfully reflect the constructs being measured.

A recent and related development is the use of multiple models as evaluators. \citeA{fonteles_novel_2026} compare multiple LLMs for multimodal classroom analysis, and \citeA{zhu_synergistic_2026} incorporate LLM-enriched Q-matrices into cognitive diagnosis. \citeA{thapa_magar_understanding_2026}, however, show that generative models can produce coherent representations of learning strategies that diverge from the actual patterns in authentic learner data; and other studies employ LLM-generated labels or evaluations without human validation \cite{misiejuk2025mapping}. 

\subsection{Detecting Student Failure Modes from Discourse}

Prior discourse-based learning analytics has used student interaction data to characterize learning-relevant states and processes, but diagnosing the specific reasoning error underlying an individual incorrect response remains comparatively underexplored. \citeA{suraworachet2024predicting} detect cognitive, metacognitive, emotional, and technical challenge moments from collaborative discourse; \citeA{booth_human-tutor_2024} analyze accountable talk in tutoring; \citeA{liu_talking_2026} connect instructional discourse with engagement and performance; \citeA{snyder_analyzing_2024} and  \citeA{zhang_using_2026} detect socially shared regulation processes; and \citeA{gures_structuring_2026} examine diagnostic reasoning and cognitive bias. These studies operate primarily at the level of broader behavioral, interactional, or challenge constructs rather than diagnosing the specific failure that resulted in an incorrect response. Authentic spoken tutoring dialogue makes this finer-grained inference difficult, as such discourse is often disfluent, interrupted, incomplete, and affected by transcription and diarization errors. Distinguishing specific failure modes may therefore require inference about learner reasoning that is only indirectly expressed in the dialogue. 

\vspace{\baselineskip}

Together, this leaves an important question unresolved for learning analytics: whether agreement across LLM families provides evidence of valid discourse interpretation, or whether shared model biases produce consensus around the same error. This has been underexplored for fine-grained diagnosis of student difficulty from authentic tutoring dialogue, including which learner states LLMs systematically over- or under-attribute and whether those biases recur across model families.

\subsection{Present Study and Contributions}

In this exploratory study, we examine whether agreement among LLMs from three contemporary model families provides a valid basis for labeling student failure modes in authentic K--12 mathematics tutoring dialogue. We study five failure modes (see Section \ref{subsec:platform_taxonomy})---uncertainty, misattribution, operator selection, conceptual gap, and procedural slip---derived from an operational codebook used by Third Space Learning\footnote{\url{https://thirdspacelearning.com/}} supporting over 196,000 students across the United States and United Kingdom. We compare labels from three models (one per family) with an adjudicated human reference set, quantify human--LLM and cross-model agreement, and analyze where and how model errors occur. We find a striking divergence between consensus and validity: while LLMs agree strongly with each other ($\kappa=.755$–$.781$), their agreement with the human consensus is substantially lower ($\kappa=.524$–$.597$), with models prone to over-inferring conceptual gaps across failure modes.

These results contribute to learning analytics in three ways. First, they provide empirical evidence that cross-model consensus should not substitute for human validation when LLMs infer complex learner constructs. Second, they identify systematic failure patterns that reveal where automated interpretation of student discourse is most vulnerable, with implications for LLM-generated labels in operational learning systems. Third, they foreground construct validity as a methodological requirement for scalable LLM-assisted annotation, showing that high agreement among models can reflect shared misinterpretation rather than faithful inference about learner reasoning. In learning analytics, consensus is not sufficient evidence of validity---particularly when the resulting labels may shape how real students are understood and supported.



\section{Methods} \label{sec:methods}

This study examines three aspects of LLM-based failure-mode diagnosis: (1) alignment with human judgments; (2) consistency across models; and (3) the failure modes and error patterns that account for the largest divergences from human interpretation. Practitioner reactions appear in the Discussion as an informal validity check, not a formal finding.

\subsection{Tutoring Platform and Failure-Mode Taxonomy} \label{subsec:platform_taxonomy}

Dialogue data are drawn from a Third Space Learning dataset of transcribed tutor--student interactions from a K--12 mathematics tutoring platform comprising 22,821 tutoring sessions (each with one or more learning objectives, or LOs) and 35,072 graded student responses (one per LO) between November 2024 and July 2025. Tutoring sessions were conducted virtually, with a tutor and student working synchronously through a shared screen on mathematics problems and discussing the student's reasoning; the resulting dialogue was recorded via the tutoring platform, then transcribed and diarized into tutor and student turns using Deepgram. Each LO is automatically marked complete once the tutor is satisfied with the student's understanding and advances to the next LO; a session's final LO is manually marked complete or incomplete by the tutor at the session's close. The corpus captures naturally occurring instructional dialogue rather than responses elicited solely for research. 

The study's five failure-mode classes are operational analytic constructs derived from the platform's diagnostic codebook and applied through an ordered decision procedure (Table~\ref{tab:taxonomy}), with a sixth label, \emph{not applicable}, used when an item cannot be classified due to insufficient information in the dialogue. The platform supports more than 196,000 students across over 4,200 schools in the United States and United Kingdom and has delivered 2.29 million tutoring sessions.

\begin{table}[h]
    \centering
    \small
    \caption{Failure-mode taxonomy, checked in the order shown; the first matching class is taken.}
    \label{tab:taxonomy}
    \begin{tabular}{@{}llp{8cm}p{5cm}@{}}
        \toprule
        \textbf{Code} & \textbf{Failure Mode} & \textbf{Definition} & \textbf{Example} \\
        \midrule
        UNC & Uncertainty & Student expresses uncertainty or reports a technical issue & ``I don't know''; ``Um...'' \\
        MIS & Misattribution & Student uses numbers or context from a different problem & Answers another question than asked\\
        OP & Operator Selection & Student chooses the wrong type of operation for this problem & Multiplies when division is needed \\
        CON & Conceptual Gap & Reasoning reveals a wrong underlying mathematical idea & Believes $0 \times n = n$ \\
        PROC & Procedural Slip & Correct strategy, wrong execution & Correct method with arithmetic error \\
        \midrule
        NA & Not Applicable & Exclusion flag (not a failure mode)---item cannot be diagnosed & Inaudible response; lack of information \\
        \bottomrule
    \end{tabular}
\end{table}

This study is a secondary analysis of de-identified, third-party tutoring transcripts. The transcript analysis was conducted in accordance with Cornell University's institutional review board requirements. Our findings concern the validity of LLM-based failure-mode classification rather than downstream learning outcomes. Because invalid labels can distort downstream learning analytics, establishing measurement validity is a prerequisite to their use.

\subsection{Data Processing and Validation}

We randomly sampled 5,000 de-identified transcripts containing student responses tied to an incomplete LO, localized to that learning-objective segment within the session (termed ``localized segments''). Localization (i.e., identifying the specific portion of the transcript containing the incorrect student response) used an LLM-based pass (Claude Sonnet 5) to identify the tutor's setup question, the corresponding graded student response, and, where present, the tutor's subsequent confirmation or correction within each learning-objective segment; this graded response served as the target for failure-mode classification. Data-quality screening removed 1,146 sampled error moments (22.9\%), primarily due to diarization error or unreliable localization. The resulting 3,854-item pool consists of graded responses, each embedded in its own localized segment, for which failure-mode localization and speaker-attribution checks passed.

Of these, 330 were randomly sampled and reserved for iterative codebook and prompt refinement, including revisions to codebook decision rules and few-shot examples. Following a human-in-the-loop, error-analysis-driven prompt engineering procedure \cite{cohn_cotal_2025}, the codebook and prompt were iteratively refined over multiple rounds ($\approx 10$), with revisions motivated by review of disagreements between LLM predictions and human judgments. Each revision was evaluated on newly drawn, disjoint holdouts rather than items used to motivate the change, reducing the risk of overfitting the annotation scheme or prompt to previously observed cases. None of these 330 development items appeared in the final adjudicated human reference set (discussed shortly) used for analysis in this paper, preserving a fully disjoint evaluation sample. The \href{https://github.com/claytoncohn/LAK27_Supplementary_Materials}{Supplementary Materials}\footnote{\url{https://github.com/claytoncohn/LAK27_Supplementary_Materials}} document the resulting decision rules, worked examples, and refinement process in full.

The remaining 3,524 items formed the sampling pool for the dataset used in this study. A preliminary LLM pass generated provisional failure-mode labels used solely for stratified sampling. We selected \textbf{$n=100$} localized segments for our adjudicated human reference set, intentionally over-sampling \emph{Conceptual Gap}, \emph{Procedural Slip}, and \emph{Operator Selection}, which the preceding refinement process had identified as particularly difficult to distinguish. Because the reference set was designed to diagnose classification failures rather than estimate population prevalence, we prioritized representation of these challenging classes over preserving their natural corpus distribution. These provisional labels were used only for sample construction and therefore do not represent the final adjudicated class distribution (presented shortly).

We chose 100 transcripts as a bounded diagnostic reference set that remained feasible for full independent annotation and adjudication while supporting overall agreement estimates and exploratory class-level error analysis. This feasibility constraint was informed by the preceding codebook and prompt refinement process, during which two researchers familiar with the codebook collectively labeled more than 500 instances across overlapping subsets of the 330 development items over two weeks. They then independently applied the finalized codebook to the graded response in each of the 100 reference-set items (i.e., localized segments), recording a label and written justification. The dialogue context available for each reference-set item ranged from 2 to 133 turns ($M=7.7$).

The two raters completed their initial segment annotations independently before discussing disagreements. Initial inter-rater agreement was modest (Cohen's $\kappa=.588$), with 29 items disputed. Following \citeA{thomas2026modernizing}, we treated these disagreements as evidence about the analytic construct rather than annotation error, resolving disagreements through adjudication against the transcript and codebook. Adjudication was completed without access to the LLM classifications, preserving the independence of the reference labels from the model predictions, and required $\approx 90$ minutes for the 29 disputed items. The resulting reference set comprised 45 \emph{Conceptual Gap} (CON), 21 \emph{Not Applicable} (NA), 14 \emph{Uncertainty} (UNC), 8 \emph{Misattribution} (MIS), 6 \emph{Procedural Slip} (PROC), and 6 \emph{Operator Selection} (OP) items. The final human-adjudicated distribution differs from the provisional LLM-based sampling allocation, including a greater number of items ultimately assigned to \emph{Conceptual Gap}. 

To contextualize our findings, we conducted semi-structured interviews (35--45 minutes) with three K--12 teachers and tutors (37, 16, and 12 years of teaching experience), who together teach (or have taught) a range of subjects including primary school mathematics, AP calculus, and language arts. Interviewees were presented the study's findings via a slide deck and asked how these model errors might affect classroom practice and what risks might follow from providing the wrong type of feedback. This procedure was approved under an IRB exemption covering the collection and use of de-identified teacher and tutor quotations. Tutor interviews were recorded via Zoom under the IRB exemption described above and transcribed using Zoom's automated transcription service. Transcripts were de-identified prior to analysis. The first author then reviewed the de-identified transcripts qualitatively to identify recurring perspectives, which are reported alongside representative quotations in Section \ref{sec:discussion}. All study artifacts, including the final annotation codebook, LLM prompts, the tutor interview slide deck, and additional procedural details, are provided in the Supplementary Materials.

\subsection{LLM Classification}

Three LLMs, one from each of three major vendors, were evaluated via API: OpenAI's GPT-5.6~Terra, Google's Gemini~3.5~Flash, and Anthropic's Claude~Opus~5. Each model was selected from the corresponding vendor's second-tier (termed ``tier 2'') stable offering to support a comparable cross-vendor evaluation while keeping inference costs tractable; deliberately spanning vendors reduces (but does not eliminate) the risk that high LLM--LLM agreement merely reflects shared training data or architecture rather than genuine interpretive convergence. Each tier 2 model's default reasoning setting (medium effort) was used. As a supplementary robustness check, we repeated the classification using Google's Gemini~3.6~Flash (the highest-performing vendor's top-performing model as of August 2026; termed ``tier 1'') with the maximum available reasoning effort ($\approx$ 65k tokens) to assess whether increased model and reasoning capability altered the observed agreement pattern. 

For all LLM runs, classifications used schema-constrained structured JSON outputs restricted to the six allowable labels and a three-run self-consistency majority vote per item. Each model received the full codebook, learning objective, task instructions, and all preceding dialogue within the localized segment up to the utterance in question. The fixed prompt included 17 expert-verified few-shot examples drawn from the development set and disjoint from the 100-item reference set. The codebook, prompts, few-shot examples, and voting procedure were then held constant across models.

\subsection{Evaluation Metrics}

Pairwise agreement between human and LLM raters and among LLMs is quantified using Cohen's $\kappa$. We use $\kappa=.70$ as a pre-specified interpretive benchmark, following its use in prior learning analytics work on automated discourse annotation \cite{liu2025qualitative}; it is not treated as a universal threshold for validity or acceptance. Multi-rater agreement across all LLMs is assessed using Krippendorff's $\alpha$. For both statistics, \emph{Not Applicable} (NA) is retained as a regular label (rather than excluded) because determining that the available evidence is insufficient to support any failure-mode diagnosis is itself a meaningful classification decision. Excluding NA cases would remove the instances needed to evaluate whether models appropriately refrain from diagnosing a failure mode when the transcript does not support one.

Per-class precision, recall, and macro-F1 identify which failure modes contribute most to LLM--human disagreement, aggregated across the three primary models. McNemar's tests evaluate whether models and human raters differ systematically in their use of NA. Item-level bootstrap resampling (1,000 resamples) provides 95\% confidence intervals for agreement estimates.

\begin{figure}[htbp]
    \centering
    \includegraphics[width=0.95\linewidth]{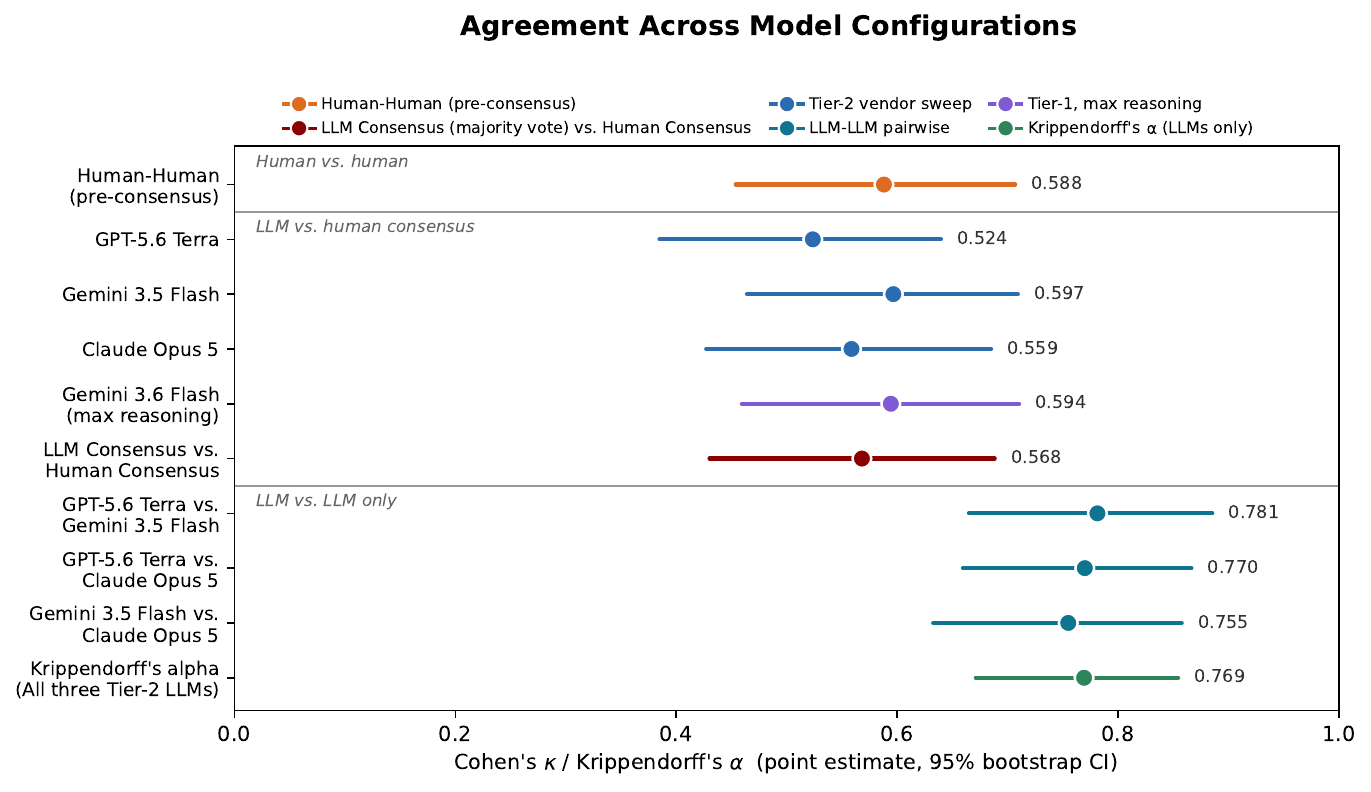}
    \caption{Agreement estimates across configurations, split into three blocks: human raters' pre-consensus agreement (top), each model's agreement with the human consensus (middle), and LLM--LLM agreement only (bottom). $\alpha$ replaces pairwise $\kappa$ for the three-LLM comparison since it generalizes to more than two raters and is on the same 0--1 scale. The top two blocks cluster closely together, while the bottom block stands well apart---illustrating the central divergence this study examines.}
    \label{fig:forestplot}
\end{figure}

\section{Findings}

\subsection{Human--LLM Agreement}

Gemini~3.5~Flash achieved the highest agreement with the human consensus ($\kappa=.597$, 95\% CI [.464, .709]), followed by Claude~Opus~5 ($\kappa=.559$, [.427, .685]) and GPT-5.6~Terra ($\kappa=.524$, [.385, .640]; Figure \ref{fig:forestplot}, middle). The tier-1 robustness check with Gemini~3.6~Flash at maximum reasoning achieved $\kappa=.594$ (95\% CI [.459, .711]), effectively matching Gemini~3.5~Flash. The human--LLM agreement gap persisted across vendors, model tiers, and reasoning settings, suggesting that it is not readily resolved by increasing model capability alone.

No model's point estimate reached the pre-specified $\kappa=.70$ benchmark: estimates ranged from .524 to .597 across the three primary models and remained at .594 for the higher-capability robustness condition, with substantially overlapping confidence intervals throughout. This LLM--human agreement was comparable in magnitude to the two human raters' initial agreement ($\kappa=.588$; Figure \ref{fig:forestplot}, top), but contrasts sharply with inter-LLM agreement. We examine this divergence next.

\subsection{LLM--LLM Agreement}

Pairwise agreement among the three tier-2 LLMs (Figure~\ref{fig:forestplot}, bottom) was consistently higher than agreement between any individual model and the adjudicated human reference set. GPT-5.6~Terra and Gemini~3.5~Flash achieved $\kappa=.781$, GPT-5.6~Terra and Claude~Opus~5 $\kappa=.770$, and Gemini~3.5~Flash and Claude~Opus~5 $\kappa=.755$. Agreement across all three LLMs was similarly high, with Krippendorff's $\alpha=.769$ (95\% CI [.671, .854]). All LLM--LLM point estimates were substantially higher than any individual model's agreement with the adjudicated human reference set, with slight confidence interval overlap.

\begin{wrapfigure}[21]{l}{0.48\textwidth}
    \centering
    \includegraphics[width=\linewidth]{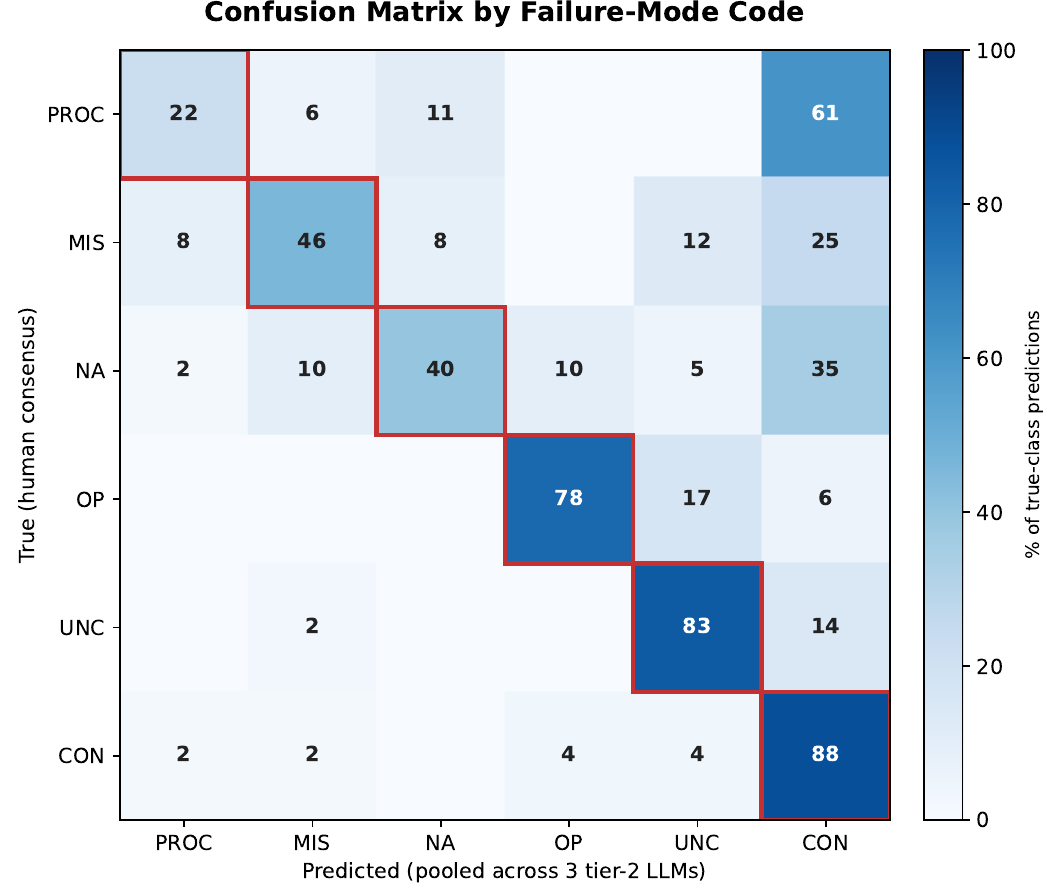}
    \caption{True (human consensus) vs.\ predicted class, pooled across the three tier-2 LLMs, row-normalized.}
    \label{fig:confusion}
\end{wrapfigure}

This divergence constitutes the study's central finding: the three models agree strongly with one another ($\kappa=.755$--$.781$; $\alpha=.769$)---conventionally interpreted as \emph{substantial} agreement \cite{landis1977measurement}---despite only \emph{moderate} agreement with the human consensus ($\kappa=.524$--$.597$) on the same 100 items. Pooling the three models into a majority-vote label (i.e., the LLM consensus; \citeA{fonteles_analyzing_2026}) did not recover the human interpretation either: the LLM-consensus label agreed with the reference set at $\kappa=.568$ (95\% CI [.430, .688]), no higher than the best individual model (Gemini~3.5~Flash, $\kappa=.597$). Cross-model consensus therefore does not indicate convergence on the human interpretation; here, it coexists with systematic disagreement about the inferred learner construct. 

LLM--LLM agreement exceeds the two human raters' pre-adjudication agreement by .17--.19 kappa points, consistent with shared systematic bias: the models exhibit the same directional tendency in their class-level errors. Taken alone, the LLM--LLM agreement in Figure~\ref{fig:forestplot} could be mistaken for evidence of a reliable classification pipeline; relative to the human reference, it instead shows how cross-model consensus can overstate validity when models share the same errors. In practice, ensemble agreement should therefore be treated as evidence of consistency, not as a substitute for human validation of LLM labels.

\subsection{LLM Error Patterns} \label{subsec:error_patterns}

This shared bias is not random: it converges on a single class, \emph{Conceptual Gap} (CON). Figure~\ref{fig:confusion} illustrates this. For every non-CON row, the largest off-diagonal cell falls in the CON column. The matrix also reveals cross-class leakage beyond CON---for example, 10\% of true NA instances are labeled MIS and another 10\% are labeled OP.

\begin{wrapfigure}[20]{r}{0.78\textwidth}
    \centering
    \includegraphics[width=\linewidth]{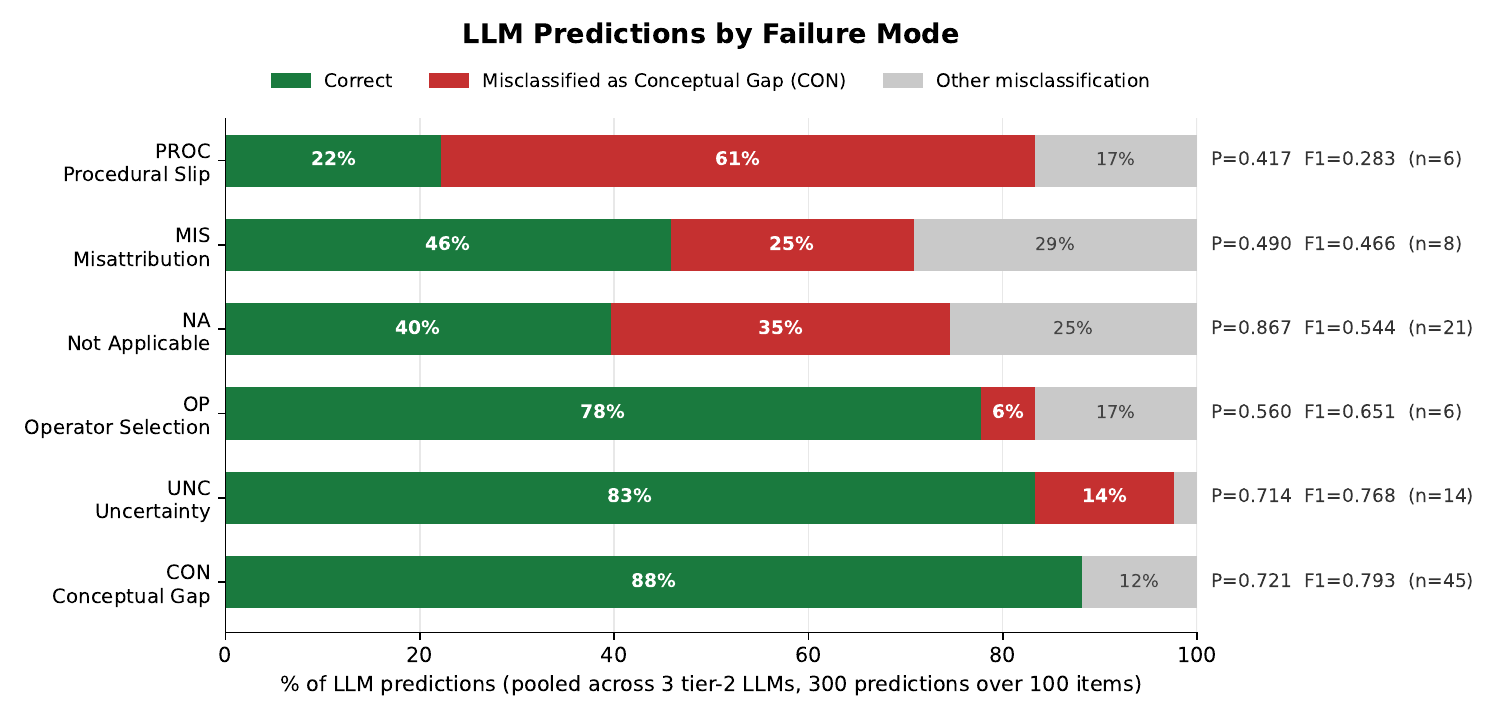}
    \caption{Pooled tier-2 LLM predictions by reference class, ordered by macro-F1. Green indicates recall; P denotes precision. For every non-CON class, the most common error is misclassification as \emph{Conceptual Gap} (red), indicating a shared directional bias.}
    \label{fig:perclass}
\end{wrapfigure}

Pooling predictions from the three tier-2 LLMs reveals the same pattern in aggregate (Figure \ref{fig:perclass}): for every non-CON class, the most common misclassification is \emph{Conceptual Gap}. The effect is strongest for \emph{Procedural Slip}, with 22\% recall; 61\% of true PROC instances are labeled CON, yielding the lowest class-level F1 ($F1=.283$, $n=6$). Because PROC and OP each contain only six human-reference items, these estimates should be interpreted as diagnostic evidence of error patterns rather than precise class performance estimates.

\emph{Not Applicable} (NA) exhibits a different but related failure pattern. NA predictions are relatively precise ($P=.867$) but have low recall (40\%), with 35\% of true NA instances instead classified as CON. All three LLMs assign NA substantially less often than the adjudicated human reference set (9--10\% vs. 21\% of items), with McNemar's tests indicating statistically significant differences for each tier-2 LLM ($p \leq .0098$). In practice, the models more often infer a conceptual difficulty than conclude that the available evidence is insufficient for diagnosis. In one such item, the student correctly stated two intermediate division facts before the response was cut off mid-step (``6 divided by 2 is 3. 6 divided by 8---so we carry the 2...[unclear]''), leaving the final answer unrecoverable from the transcript; the human reference therefore withheld a diagnosis. All three LLMs nonetheless converged on \emph{Conceptual Gap}, illustrating consensus despite insufficient evidence for any failure-mode diagnosis.

Together, these patterns help explain the divergence between human--LLM and LLM--LLM agreement: models systematically over-attribute student difficulty to conceptual deficits, both when another failure mode is more appropriate and when no diagnosis is supported. These errors represent a shared directional bias rather than random noise that might average out across models. For learning analytics, such bias can systematically distort representations of learner difficulty.

\section{Discussion and Conclusions} \label{sec:discussion}

This study demonstrates that cross-model agreement is not necessarily evidence of valid learner interpretation. Across three LLM families, models agreed more with each other than with the human consensus while sharing systematic directional errors. For learning analytics, LLM consensus can make annotation appear reliable while masking shared misinterpretation.

\subsection{Validity of Learner Representations in Learning Analytics}

Our findings point to a broader measurement problem in learning analytics: agreement among annotators, human or model-based, does not establish that learner representations validly capture the underlying construct. This is consistent with \citeA{thomas2026modernizing}, who argue that inter-rater reliability should inform examination of disagreement and construct refinement rather than serve as a mechanical gatekeeper for ground truth. Similarly, \citeA{henkel2025humans} found that models remained highly consistent in formative literacy assessment even as model--human agreement declined. In our case, LLMs produced consistent representations of student difficulty that diverged substantially from human consensus.

\subsection{Instructional Consequences of Diagnostic Misclassification}

When diagnostic labels drive instructional responses, misclassification has direct pedagogical consequences. Automated tutoring systems use inferred error types to select targeted feedback and remediation \cite{mcnichols2023algebra,reddig2025generating}. Misclassifying a \emph{Procedural Slip} as a \emph{Conceptual Gap}, for example, can prompt feedback that treats an execution error as deficient understanding, despite evidence that responses to student errors should support motivation and build on what the learner actually knows \cite{thomas2023tutor}. One interviewed tutor described the risk of internalized harm from repeated misdirected correction: ``even if it's that procedural slip... they're gonna start to internalize some of that, like, I'm bad at math.'' 

Conversely, misclassifying a genuine \emph{Conceptual Gap} as a procedural slip can leave the underlying misconception insufficiently addressed, allowing it to persist \cite{gurung2023identification}. Diagnostic errors are therefore not interchangeable: different misclassifications can produce different instructional consequences---a point interviewed tutors raised independently of the pedagogical literature. As one observed, reteaching a skill that was never actually missing wastes instructional time: ``if you're reteaching something because something has been misdiagnosed as something else... you're not moving forward...what was supposed to be a tool that helps teachers is now causing more pain and time.'' Learning analytics systems should evaluate not only whether learner states are classified accurately, but also the consequences of acting on incorrect diagnoses.

\subsection{Rethinking Traditional Discourse Annotation}

The validity problem is not exclusively computational: disagreement among human annotators indicates that the underlying learner construct is itself difficult to infer from transcript evidence alone. This suggests that mutually exclusive failure-mode labels may impose sharper boundaries than the data support. Multi-label and uncertainty-aware annotation may better preserve cases in which several interpretations remain plausible.

From an ICAP perspective, these exchanges provide observable evidence of constructive and interactive engagement, but our study highlights a complementary problem: even when discourse contains learning-relevant evidence, transforming that evidence into a diagnostic learner representation can introduce systematic measurement error. Transcript-only data also omit information available during tutoring, including prosody, written work, screen state, gestures, and other interaction traces. Some disagreement may therefore reflect missing evidence rather than deficient reasoning. Prior learning analytics work demonstrates the value of combining discourse with complementary trace data for contextualization \cite{snyder_analyzing_2024}, suggesting valid failure-mode diagnosis may benefit from multimodal evidence.

\subsection{Limitations and Future Work}

This exploratory study uses 100 deliberately sampled error moments from a single K--12 mathematics tutoring platform, limiting population-level and cross-domain generalizability; several classes also remain small. Practitioner perspectives were drawn from semi-structured discussions with three tutors rather than a systematic or representative stakeholder sample, and student perspectives were not collected. We also evaluate diagnostic validity rather than downstream consequences: the effects of failure mode misclassification on feedback, self-efficacy, misconception persistence, engagement, and learning remain unknown and should be examined alongside broader stakeholder perspectives. Future work should also consider larger reference sets, multimodal evidence, multi-label taxonomies, and uncertainty-aware inference; and should assess whether these findings generalize across platforms, subjects, and tutor and student populations.

The central implication is nevertheless clear: before learning analytics asks whether LLM-generated diagnoses improve learning, it must establish that those diagnoses validly represent learner difficulty. When models share the same biases, consensus can increase confidence without increasing correctness. \emph{Agreement demonstrates consistency; validity requires evidence that the learner has been interpreted correctly.}









\phantomsection
\bibliography{references}


\end{document}